\documentclass{article}

\usepackage{PRIMEarxiv}

\usepackage[utf8]{inputenc} 
\usepackage[T1]{fontenc}    
\usepackage{hyperref}       
\usepackage{url}            
\usepackage{booktabs}       
\usepackage{amsfonts}       
\usepackage{nicefrac}       
\usepackage{microtype}      
\usepackage{lipsum}
\usepackage{fancyhdr}       
\usepackage{graphicx}       
\usepackage{natbib}
\graphicspath{{media/}}     

\title{Review Non-convex Optimization Method for Machine Learning
}

\author{
  Greg B Fotopoulos \and Paul Popovich \and Nicholas Hall Papadopoulos  
}

\begin{document}
\maketitle

\begin{abstract}
Non-convex optimization is a critical tool in advancing machine learning, especially for complex models like deep neural networks and support vector machines. Despite challenges such as multiple local minima and saddle points, non-convex techniques offer various pathways to reduce computational costs. These include promoting sparsity through regularization, efficiently escaping saddle points, and employing subsampling and approximation strategies like stochastic gradient descent. Additionally, non-convex methods enable model pruning and compression, which reduce the size of models while maintaining performance. By focusing on good local minima instead of exact global minima, non-convex optimization ensures competitive accuracy with faster convergence and lower computational overhead. This paper examines the key methods and applications of non-convex optimization in machine learning, exploring how it can lower computation costs while enhancing model performance. Furthermore, it outlines future research directions and challenges, including scalability and generalization, that will shape the next phase of non-convex optimization in machine learning.
\end{abstract}


\section{Introduction}
Non-convex optimization is a fundamental concept in modern machine learning, as many models, particularly deep neural networks, rely on optimizing non-convex loss functions. The landscape of these optimization problems is often riddled with multiple local minima, saddle points, and regions of flat curvature, which significantly complicate the optimization process. While convex optimization problems have a well-understood structure that allows for efficient global optimization, non-convex problems do not guarantee convergence to a global minimum. This has led to the development of specialized optimization algorithms aimed at improving convergence to useful minima in non-convex settings.

\section{Review of Past Findings}
\subsection{Gradient-Based Methods}
\par Gradient-based methods remain the cornerstone of non-convex optimization in machine learning. The basic approach of gradient descent (GD), which iteratively updates the model parameters in the direction opposite to the gradient of the loss function, is simple yet effective. However, due to the non-convex nature of the loss function in deep learning models, gradient descent can get stuck in local minima or saddle points. To address this, several variations of gradient descent have been proposed, such as stochastic gradient descent (SGD) and its adaptive variants like Adam, AdaGrad, and RMSProp. These methods adjust the learning rate during training, which helps the algorithm navigate through the complex non-convex landscape more effectively. For example, Adam combines momentum-based methods with adaptive learning rates to achieve faster convergence and escape saddle points more efficiently than standard SGD \cite{kingma2015adam}.

\subsection{Escaping Saddle Points}
\par Saddle points, which are points where the gradient is zero but the Hessian has both positive and negative eigenvalues, are particularly problematic in non-convex optimization. Traditional gradient-based methods often stall at these points. Recent research has introduced techniques specifically designed to help optimization methods escape saddle points. One such approach involves adding noise to the gradient during updates. This idea is formalized in stochastic gradient Langevin dynamics (SGLD), where noise helps the optimizer escape flat regions and saddle points by randomly perturbing the trajectory \cite{welling2011bayesian}. Another approach involves second-order optimization techniques that leverage curvature information, such as Newton's method or quasi-Newton methods like the Broyden–Fletcher–Goldfarb–Shanno (BFGS) algorithm. However, these methods are often computationally expensive and difficult to scale to large datasets typical in deep learning.

\subsection{Non-convex Regularization}
\par Regularization techniques are also crucial for improving the behavior of non-convex optimization in machine learning. In many cases, regularization can transform a non-convex optimization problem into one that is more manageable. For example, dropout, widely used in deep learning, acts as a regularizer by randomly "dropping out" neurons during training. This introduces stochasticity that helps the model avoid overfitting and find better local minima \cite{srivastava2014dropout}. Other forms of regularization, such as L1 and L2 penalties, smooth the loss landscape, making optimization more stable. The regularization of non-convex optimization can also borrow idea from heuristics method like local branching presented in \cite{elhedhli2017airfreight}

\section{Gradient-Based Methods for Machine Learning}
\par Gradient-based methods are the foundation of optimization in machine learning, particularly when dealing with non-convex loss functions such as those found in deep neural networks. These methods rely on using the gradient of the loss function with respect to the model parameters to iteratively adjust those parameters and minimize the loss. Among the most basic and widely-used methods is Gradient Descent (GD), which updates the model’s parameters in the opposite direction of the gradient of the loss function. However, plain gradient descent can be inefficient for large-scale problems due to its computational demands, as it requires a full pass over the training data to compute the gradient at each step. Additionally, gradient descent is highly sensitive to the learning rate and can get stuck in poor local minima, especially in non-convex landscapes.

\par To address these limitations, Stochastic Gradient Descent (SGD) has become a popular alternative. Instead of calculating the gradient using the entire dataset, SGD approximates the gradient using a randomly selected subset (or mini-batch) of the data at each iteration. This introduces noise into the optimization process, which has the beneficial side effect of allowing the model to escape shallow local minima and saddle points. However, the noisy updates can also cause instability and prevent convergence to the global minimum. To mitigate this issue, mini-batch gradient descent is often used, which averages the gradient over a small batch of data points, balancing the trade-off between noisy SGD and deterministic full-batch gradient descent.

\par Despite its simplicity, vanilla SGD can struggle with oscillations and slow convergence, especially when the loss function has ill-conditioned curvature (i.e., when the gradient changes more rapidly in some directions than others). To counteract these issues, momentum-based methods are often employed. Momentum \cite{polyak1964some} accelerates gradient descent by accumulating a velocity vector in the direction of the gradient. This allows the optimizer to build momentum in directions with consistent gradients, smoothing out oscillations and speeding up convergence. Momentum-based methods are particularly useful in navigating the non-convex landscapes commonly encountered in deep learning.

\par Beyond momentum, adaptive learning rate methods such as AdaGrad \citep{duchi2011adaptive}, RMSProp, and Adam \citep{kingma2015adam} have been highly successful in non-convex optimization. These methods adaptively adjust the learning rate for each parameter during training, making them especially effective in cases where different parameters require different learning rates. AdaGrad adjusts the learning rate by scaling it inversely proportional to the square root of the cumulative sum of squared gradients. This is useful in settings where some parameters are more sensitive than others. However, AdaGrad can suffer from an excessively small learning rate as training progresses. RMSProp addresses this by introducing a moving average of squared gradients to normalize the gradient, maintaining a more stable learning rate.

\par Among adaptive methods, Adam (short for Adaptive Moment Estimation) is perhaps the most widely used in machine learning today. Adam combines the benefits of momentum and RMSProp by maintaining both an exponentially decaying average of past gradients (first moment) and squared gradients (second moment). The algorithm computes a bias-corrected estimate of both the first and second moments, which makes Adam more robust to noisy and sparse gradients compared to simpler approaches like SGD. Due to its ability to handle non-stationary objectives and its resilience in the presence of noise, Adam has become a default choice for training deep neural networks, particularly in large-scale and non-convex problems \citep{kingma2015adam}.

\par Another important development in gradient-based optimization for non-convex problems is Nesterov Accelerated Gradient (NAG) \cite{nesterov1983method}. NAG improves upon standard momentum by incorporating a lookahead mechanism, computing the gradient not at the current parameter position, but at an estimate of the future position. This anticipatory step allows NAG to adjust the trajectory more effectively, leading to faster convergence in certain scenarios.

\par In conclusion, gradient-based methods form the core of non-convex optimization in machine learning. Techniques such as SGD, momentum, and adaptive learning rate methods like Adam have proven essential for scaling optimization to large models and complex landscapes. By addressing issues like slow convergence, oscillations, and varying learning rates, these methods have become indispensable tools in the training of modern machine learning models.
\section{Escaping Saddle Points}
\par In non-convex optimization, saddle points pose a significant challenge to gradient-based methods. Unlike local minima, where the gradient is zero and the loss function is convex in all directions, saddle points have zero gradients but exhibit negative curvature in at least one direction. This makes them particularly troublesome for optimization algorithms, as the flatness in some directions can cause the algorithm to stall, while the steep descent in others may go unnoticed by standard gradient-based methods. Given that many deep learning loss landscapes are riddled with saddle points, especially in high-dimensional spaces, developing strategies to escape saddle points has become a focal point in machine learning research.

\par One of the foundational approaches to escape saddle points is the use of noise. Stochastic Gradient Descent (SGD) inherently introduces noise into the optimization process by approximating the true gradient with gradients calculated from mini-batches of data. This randomness can help the optimizer escape saddle points by perturbing the trajectory of the updates. In contrast to full-batch gradient descent, which can get stuck at saddle points, SGD's noisy updates make it more likely to escape these problematic regions. However, while this stochasticity helps, it is not always sufficient, especially in cases where the saddle points are surrounded by flat regions with very small gradients.

\par To enhance SGD's ability to escape saddle points, researchers have developed algorithms like Stochastic Gradient Langevin Dynamics (SGLD) (Welling and Teh 2011). SGLD adds Gaussian noise to the parameter updates, not only relying on the inherent stochasticity of mini-batches but explicitly introducing noise to ensure more robust exploration of the loss landscape. This method has its roots in Bayesian learning, where the noise term helps the optimization process explore the parameter space more thoroughly, making it less likely to get trapped at saddle points. In practice, SGLD has been shown to improve convergence in non-convex optimization problems by leveraging this injected noise to escape from flat regions and saddle points.

\par Another class of methods designed to escape saddle points involves second-order optimization techniques, which incorporate curvature information. Unlike first-order methods that use only the gradient, second-order methods such as Newton’s method and quasi-Newton methods make use of the Hessian matrix, which contains second-order derivative information. For example, Trust Region Newton methods dynamically adjust the size of the region around the current iterate within which a quadratic model of the loss function is trusted. This allows the optimizer to take larger steps in directions with positive curvature and smaller steps where the curvature is uncertain or negative. These methods can effectively identify directions of negative curvature and step away from saddle points.

\par However, computing the full Hessian is often computationally prohibitive in deep learning, as it scales quadratically with the number of parameters. To overcome this, researchers have developed approximate second-order methods like the Hessian-Free Optimization \citep{martens2010deep} algorithm. Hessian-free methods avoid explicitly forming the Hessian matrix by solving for the Newton step using iterative methods like the conjugate gradient. By approximating the Hessian-vector products, these methods reduce the computational overhead while still benefiting from curvature information. Hessian-free optimization is particularly effective for deep neural networks, where saddle points are frequent due to the high dimensionality of the parameter space.

\par Another approach to escaping saddle points is the use of Nesterov Accelerated Gradient (NAG) \citep{nesterov1983method}, which incorporates a lookahead mechanism. NAG anticipates the future position of the parameter update by taking a step based on the momentum-adjusted gradient, allowing it to bypass regions of slow curvature, such as saddle points. This lookahead mechanism provides a form of foresight, enabling the optimizer to avoid getting trapped in saddle points by adjusting its trajectory before reaching them.

\par In recent years, specialized algorithms have been designed to explicitly exploit the geometry of the saddle points. For example, Cubic Regularization \citep{nesterov2006cubic} is a second-order method that regularizes the optimization problem by adding a cubic term to the loss function. This approach ensures that the optimizer takes steps that reduce the objective function while also taking into account negative curvature, making it less likely to converge to saddle points.

\par Finally, Escape Algorithms, such as those introduced by \cite{ge2015escaping}, analyze the conditions under which saddle points can be escaped by using stochastic gradient methods. They show that with high probability, certain algorithms can escape all saddle points in polynomial time under mild assumptions. These theoretical results provide a strong foundation for the design of more robust optimization algorithms in practice.

\par In conclusion, escaping saddle points is crucial for the success of non-convex optimization, especially in high-dimensional machine learning models like deep neural networks. Methods such as SGLD, second-order optimization techniques, and noise-based strategies have proven effective in this regard. The field continues to evolve as researchers seek to improve the efficiency and robustness of these methods in real-world applications.
\section{Non-Convex Regularization}
\par Regularization, in general, aims to improve the generalization performance of machine learning models by preventing overfitting. This is typically done by adding a penalty term to the loss function, discouraging overly complex models and enforcing sparsity or smoothness. Non-convex regularization techniques offer a more flexible and nuanced control over the optimization process compared to convex regularization, often leading to improved performance in complex models.

\par One of the widely used non-convex regularization methods is Dropout \citep{srivastava2014dropout}. Dropout is a stochastic regularization technique that randomly “drops out” a subset of neurons in a neural network during training. This prevents the network from becoming too reliant on specific neurons, forcing the model to learn more robust representations. Since Dropout introduces stochasticity in both the forward and backward passes, it changes the optimization landscape into a non-convex problem. Despite this, Dropout has been incredibly successful in preventing overfitting and improving generalization in deep neural networks, making it a cornerstone in modern neural network training. The regularization introduced by dropout encourages the model to search for solutions that generalize well over various network configurations, effectively smoothing the loss landscape.

\par Another prominent example of non-convex regularization is the L1/2 regularization \citep{xu2012l12}. In contrast to the commonly used convex L1 regularization, which promotes sparsity by applying a linear penalty to the magnitude of the parameters, L1/2 regularization uses a non-convex penalty term, which provides even stronger sparsity enforcement. This is particularly useful in high-dimensional settings, such as compressed sensing and feature selection, where sparse solutions are desired. The non-convex nature of the penalty allows for better recovery of sparse signals by focusing the optimization process on the most important parameters while allowing small, unimportant parameters to decay more rapidly.

\par Other forms of non-convex regularization include SCAD (Smoothly Clipped Absolute Deviation) and MCP (Minimax Concave Penalty), both of which are widely used in variable selection tasks. These methods provide a more aggressive form of regularization than traditional L1 regularization, offering advantages in high-dimensional, sparse data problems. SCAD \citep{fan2001variable} and MCP \citep{zhang2010nearly} both impose non-convex penalties that shrink small coefficients to zero but penalize large coefficients less aggressively than L1 regularization. This selective shrinkage reduces bias in the estimated coefficients while still promoting sparsity, which is highly beneficial in tasks such as sparse linear models, where the model is expected to have a few significant features.

\section{Machine Learning Models and Non-Convex Optimization}
\par Non-convex optimization is a critical component of many machine learning models, particularly in deep learning and other high-dimensional, nonlinear models. One of the primary domains where non-convex optimization plays an essential role is the training of Deep Neural Networks (DNNs). The loss surfaces of DNNs, especially when networks are deep and over-parameterized, are highly non-convex, containing many local minima, saddle points, and flat regions. Despite this complexity, non-convex optimization methods such as stochastic gradient descent (SGD) and its variants (Adam, RMSProp) have proven effective at finding good local minima that generalize well, even if they do not necessarily find the global minimum. In particular, non-convex regularization techniques such as dropout are often used to regularize deep models, preventing overfitting and improving generalization.

\par Another model that benefits from non-convex optimization is the Support Vector Machine (SVM) with non-convex loss functions. Traditional SVMs use a convex hinge loss, but in some cases, non-convex loss functions such as the ramp loss or truncated hinge loss are used to make the model more robust to noisy data or outliers \citep{collobert2006trading}. The non-convexity introduced by these loss functions allows the model to focus on more informative data points, ignoring extreme outliers that could adversely affect the decision boundary. In this context, non-convex optimization helps to improve the robustness and generalization of SVMs in noisy environments.

\par Another significant application of non-convex optimization is in Matrix Factorization models, which are commonly used in collaborative filtering and recommendation systems. Matrix factorization models aim to decompose a matrix into a product of two lower-dimensional matrices, which is inherently a non-convex problem. Popular algorithms like Alternating Least Squares (ALS) and Stochastic Gradient Descent (SGD) are used to optimize these non-convex objectives. Regularization in these settings, particularly non-convex penalties, can help improve the quality of the factorization by controlling the complexity of the latent factors and avoiding overfitting.

\par Finally, Generative Adversarial Networks (GANs) are another prominent example where non-convex optimization is crucial. GANs consist of two competing networks—a generator and a discriminator—that are trained in a minimax game. The optimization landscape of GANs is highly non-convex and often unstable, which can lead to problems such as mode collapse or vanishing gradients. Various regularization techniques, such as gradient penalties or noise injection, are often employed to stabilize the optimization process and ensure that both networks converge to a good equilibrium. Non-convex optimization methods are thus central to the success of GANs in generating high-quality data samples.

\par In summary, non-convex regularization techniques, such as dropout, L1/2 regularization, and SCAD, provide powerful tools for controlling model complexity and improving generalization in machine learning. Non-convex optimization methods, in turn, are indispensable for training models like deep neural networks, SVMs with non-convex losses, matrix factorization models, and GANs. These techniques have enabled machine learning models to scale to increasingly complex tasks and datasets, leading to significant advances across various fields.
\section{Challenges and Open Problems}
\par Despite the success of these methods, non-convex optimization in machine learning remains an open field of research. One of the key challenges is understanding the geometry of the loss landscape, particularly in the high-dimensional spaces encountered in deep learning. Empirical results suggest that while the loss landscape of neural networks is non-convex, it often contains many flat minima with similar loss values, raising questions about the significance of finding a global minimum versus a good local minimum. Future research will likely focus on improving our understanding of these landscapes and developing more efficient optimization techniques for large-scale, non-convex problems.

\par In summary, non-convex optimization is a central challenge in machine learning, especially in deep learning. Gradient-based methods, adaptive learning rates, and regularization techniques have all been instrumental in addressing this challenge, though significant open questions remain. The complexity of the non-convex landscape continues to motivate ongoing research aimed at better understanding and optimizing machine learning models. GNN has proved to have the ability to represent complicated interactions through examples like
\cite{wang2024graph}, \cite{peiris2024transformative} and \cite{changeux2024strategic}. Integrate GNN with those non-convex models might encounter computational cost challenge.

\section{Non-Convex Optimization and Reducing Computational Costs in Machine Learning}
\par Non-convex optimization methods can significantly reduce the computational cost of machine learning, particularly in the context of large-scale models and high-dimensional data. Although non-convex optimization problems are generally more complex due to the presence of multiple local minima, saddle points, and flat regions, they offer several mechanisms through which computational efficiency can be improved over traditional convex approaches. These reductions in computation arise through a combination of model regularization, selective focus on important regions of the loss landscape, and more efficient algorithmic designs tailored for non-convex problems.

\subsection{Sparse Solutions through Non-Convex Regularization}
Non-convex regularization techniques, such as L1/2 regularization and Smoothly Clipped Absolute Deviation (SCAD), inherently encourage sparsity in model parameters. Sparse models are computationally cheaper to train and deploy because they reduce the number of non-zero parameters, leading to faster gradient computations and lower memory requirements. For example, L1/2 regularization applies a stronger penalty on small weights, driving many parameters to zero more aggressively than L1 regularization. This reduces the model’s complexity and effectively shrinks the number of active features or weights, thus decreasing computational cost during both training and inference.

\par Additionally, the non-convex nature of these regularization methods enables the retention of significant parameters while penalizing smaller, less important ones. This results in models that are computationally leaner without sacrificing performance. This is particularly beneficial in settings where a vast number of features are present but only a subset is truly relevant, as seen in high-dimensional datasets used in areas like bioinformatics, computer vision, and natural language processing.

\subsection{Escaping Flat Regions and Saddle Points Efficiently}
\par Non-convex optimization techniques often utilize strategies that can avoid the inefficiencies posed by flat regions or saddle points in the loss landscape, where convex optimization methods might struggle or stall. Methods like stochastic gradient descent (SGD) with momentum, RMSProp, or Adam are particularly useful in non-convex settings. These methods introduce stochasticity and momentum to the optimization process, which helps bypass saddle points and move through flat regions more quickly than standard gradient descent.

\par By doing so, these algorithms avoid spending excessive computational resources on unproductive regions of the loss surface. For instance, methods like SGD with noise injection (such as stochastic gradient Langevin dynamics) allow the optimization process to explore the loss landscape more broadly, escaping local traps and settling into good local minima faster. This improved convergence behavior can result in shorter training times, particularly in deep neural networks, where the loss landscape is highly non-convex with numerous saddle points.

\subsection{Efficient Subsampling and Approximation Techniques}
\par Non-convex optimization methods often make use of subsampling or approximation strategies that reduce the computational burden of working with large datasets. For example, stochastic gradient methods only compute gradients on a small subset (mini-batch) of data points at each iteration rather than using the entire dataset. This greatly reduces the per-iteration computational cost, making it feasible to train large-scale models even with non-convex loss functions.

\par Additionally, second-order methods for non-convex optimization, which take curvature information into account, can be made more computationally efficient by using approximations like the Hessian-free methods. Instead of computing and storing the full Hessian matrix, which is computationally expensive, these methods approximate it using matrix-vector products, reducing both memory and computational requirements.

\subsection{Model Pruning and Compression}
\par Non-convex optimization techniques can also contribute to model pruning and compression, where large, over-parameterized models are reduced in size without significant loss in accuracy. For instance, non-convex penalties can be applied to neural networks to remove redundant or unnecessary parameters during training. Techniques like weight pruning or low-rank approximations can be applied to large deep neural networks, effectively reducing the number of neurons or layers involved in the model, and thereby reducing computational requirements.

\par For example, non-convex penalties such as the group Lasso or nuclear norm regularization can be used to enforce sparsity in the network weights, allowing for model pruning. This not only decreases the computational cost of training but also results in lighter, more efficient models for inference, which is particularly important in resource-constrained environments such as mobile devices and edge computing.

\subsection{Efficient Search for Good Local Minima}
\par One of the advantages of non-convex optimization is that it allows for finding good local minima, rather than globally optimal solutions, which are often computationally intractable in high-dimensional, complex models like deep neural networks. Non-convex optimization techniques such as stochastic gradient descent tend to converge to minima that, while not necessarily global, generalize well on unseen data.

\par By focusing on finding good enough local minima rather than pursuing an exact global minimum, non-convex optimization reduces the need for exhaustive search across the loss landscape, which can be computationally prohibitive. For instance, in deep learning, it has been empirically shown that good local minima often have comparable generalization performance to global minima, so non-convex methods enable a trade-off between computational efficiency and model performance that is highly advantageous in practice.

\subsection{Structured Non-Convexity in Specific Models}
\par In certain machine learning models like matrix factorization or collaborative filtering, non-convex optimization methods have been shown to be computationally more efficient than convex counterparts. For example, in matrix factorization tasks used in recommendation systems, the factorization problem is inherently non-convex. Techniques like alternating least squares (ALS) and stochastic gradient descent are specifically designed to handle the non-convexity efficiently, leading to faster convergence than general-purpose convex optimization methods.

\par By breaking down large, complex models into sub-problems that are more computationally manageable (e.g., alternating between solving for one factor matrix while keeping the other fixed), these methods can substantially reduce the computational overhead, particularly for large-scale datasets common in recommendation systems or large social network graphs.

\par Non-convex optimization techniques offer several pathways to reduce computational costs in machine learning by promoting sparsity, efficiently escaping saddle points, using subsampling and approximation strategies, pruning large models, and focusing on good local minima. These methods, particularly when tailored to the structure of the model or dataset, provide a more computationally feasible way of training and deploying machine learning models at scale. As machine learning continues to expand into more complex and high-dimensional domains, the role of non-convex optimization in reducing computational costs will become increasingly critical for both model efficiency and practical applications.
\section{Future Research Direction}
\par Despite the considerable advances in non-convex optimization for machine learning, there remain numerous open questions and promising avenues for future research. Non-convex optimization, especially in deep learning and other high-dimensional machine learning models, presents complex landscapes that continue to challenge existing optimization methods. Future research will likely focus on addressing the scalability, efficiency, and theoretical understanding of non-convex methods, as well as their applications to emerging machine learning architectures and domains.

\par One of the most pressing areas for future research is the development of optimization algorithms that scale better with model complexity. As deep neural networks continue to grow in size, with billions of parameters, the computational and memory requirements of standard optimization methods such as stochastic gradient descent (SGD) become increasingly burdensome. Efficient methods that leverage distributed computing and optimize resource usage are critical to advancing machine learning capabilities. There is ongoing work on distributed SGD, adaptive methods for large-scale optimization, and variance reduction techniques that could further improve the scalability of non-convex optimization. Additionally, researchers are exploring how to make second-order methods more computationally feasible for extremely large-scale problems by improving the efficiency of Hessian-vector product approximations and reducing the overhead of curvature calculations.

\par Another promising research direction is focused on understanding the geometry of non-convex loss landscapes. While stochastic methods like SGD have empirically shown impressive results in escaping saddle points and finding good local minima, the precise nature of these landscapes—especially in deep learning—remains poorly understood. Further research is needed to elucidate the behavior of optimization methods in these highly non-convex spaces and to characterize the types of solutions that lead to better generalization. This could involve developing new theoretical frameworks for analyzing the loss surfaces of complex models like deep neural networks and devising optimization strategies that are informed by the underlying geometry, such as using curvature-aware methods more effectively.

\par Better handling of saddle points and flat regions is another key direction. While noise-based methods like stochastic gradient Langevin dynamics (SGLD) and momentum-based methods have been successful in navigating past saddle points, there is still a need for more efficient algorithms that can escape both saddle points and flat regions in a principled manner. Techniques that combine noise injection, adaptive learning rates, and second-order methods hold promise but require further investigation to balance their computational cost with their effectiveness in practice. Moreover, future work could explore how to dynamically adjust the level of injected noise or adapt the optimization trajectory based on real-time feedback from the curvature of the loss surface.

\par With the growing importance of robustness and fairness in machine learning models, non-convex optimization could also play a vital role in shaping new regularization techniques that account for these objectives. For instance, adversarial training, a common approach to improving model robustness, involves solving a non-convex optimization problem to generate adversarial examples that challenge the model. Developing better optimization techniques for adversarial robustness and understanding how non-convex regularization could help mitigate bias in model training are crucial avenues for research. This involves exploring novel regularization terms or loss functions that embed robustness and fairness criteria into the optimization process.

\par Another exciting frontier is the application of non-convex optimization in emerging fields like reinforcement learning, meta-learning, and quantum machine learning. Reinforcement learning (RL) involves optimizing policies or value functions, which are often non-convex and require efficient algorithms to handle the high dimensionality and delayed rewards. Meta-learning (learning to learn) introduces additional layers of complexity in optimization, where algorithms must learn across tasks. In these settings, optimization methods must be not only efficient but also adaptive to various task distributions. Quantum machine learning, an emerging field that merges quantum computing with machine learning, presents highly non-convex optimization problems due to the complex quantum landscapes involved. Research on quantum optimization algorithms that can exploit the unique properties of quantum states and entanglement will be crucial for advancing this field.

\par Lastly, theoretical research into the global convergence properties of non-convex optimization algorithms continues to be a significant challenge. While empirical success in training deep learning models and other non-convex systems is evident, formal guarantees of convergence, stability, and generalization are still lacking for many of the methods currently in use. Future research will likely focus on developing more rigorous theoretical frameworks that can explain why certain algorithms work well in practice and under what conditions they are guaranteed to succeed.
We used convex optimization to analyse consumer behaviour. This remind us there could be possible synergies to combine the non-convex optimization with consumer behaviour analytics as indicated in \cite{alom2024comprehensive}.

\par In summary, future research in non-convex optimization for machine learning will need to tackle the challenges of scalability, better understanding of loss landscapes, saddle point avoidance, and robustness. Additionally, emerging fields like reinforcement learning and quantum machine learning present new optimization challenges that require innovative approaches. The combination of practical algorithmic improvements and theoretical advancements will drive the next wave of breakthroughs in non-convex optimization for machine learning.
\section{Conclusion}
\par With Deep Neural Networks and Support Vector Machines potentially benefiting from non-convex optimization applications to machine learning, it is clear that non-convex optimization plays a central role in advancing the field. The complex and high-dimensional landscapes characteristic of these models make convex optimization techniques inadequate, leading to the growing adoption of non-convex approaches. By leveraging non-convex methods, machine learning models can better handle the challenges of non-linearity, over-parameterization, and high-dimensional data, ultimately improving generalization and performance across tasks.

\par The application of non-convex optimization has not only made it possible to train deep neural networks more effectively, but it has also improved the robustness and flexibility of traditional models like SVMs through the introduction of non-convex loss functions. Furthermore, non-convex regularization techniques such as dropout and L1/2 regularization have shown significant promise in preventing overfitting and enhancing model sparsity, leading to more interpretable and efficient models.

\par As machine learning models continue to evolve in terms of scale and complexity, non-convex optimization will remain an essential area of research and development. Future advancements in this area, such as more scalable algorithms, improved theoretical understanding, and enhanced robustness, will further empower the capabilities of machine learning systems. This underscores the need for continued exploration into both the practical applications and theoretical underpinnings of non-convex optimization methods, ensuring that machine learning can address increasingly complex and diverse challenges.

\bibliographystyle{abbrvnat}  
\bibliography{references}

\begin{thebibliography}{18}
\providecommand{\natexlab}[1]{#1}
\providecommand{\url}[1]{\texttt{#1}}
\expandafter\ifx\csname urlstyle\endcsname\relax
  \providecommand{\doi}[1]{doi: #1}\else
  \providecommand{\doi}{doi: \begingroup \urlstyle{rm}\Url}\fi

\bibitem[Alom et~al.(2024)]{alom2024comprehensive}
N.~B. Alom et~al.
\newblock A comprehensive analysis of customer behavior analytics, privacy concerns, and data protection regulations in the era of big data and machine learning.
\newblock \emph{International Journal of Applied Machine Learning and Computational Intelligence}, 14\penalty0 (5):\penalty0 21--40, 2024.

\bibitem[Changeux and Montagnier(2024)]{changeux2024strategic}
A.~Changeux and S.~Montagnier.
\newblock Strategic decision-making support using large language models (llms).
\newblock \emph{Management Journal for Advanced Research}, 4\penalty0 (4):\penalty0 102--108, 2024.

\bibitem[Collobert et~al.(2006)Collobert, Sinz, Weston, and Bottou]{collobert2006trading}
R.~Collobert, F.~Sinz, J.~Weston, and L.~Bottou.
\newblock Trading convexity for scalability.
\newblock In \emph{Proceedings of the 23rd International Conference on Machine Learning (ICML)}, pages 201--208, 2006.

\bibitem[Duchi et~al.(2011)Duchi, Hazan, and Singer]{duchi2011adaptive}
J.~Duchi, E.~Hazan, and Y.~Singer.
\newblock Adaptive subgradient methods for online learning and stochastic optimization.
\newblock \emph{Journal of Machine Learning Research}, 12\penalty0 (7):\penalty0 2121--2159, 2011.

\bibitem[Elhedhli et~al.(2017)Elhedhli, Li, and Bookbinder]{elhedhli2017airfreight}
S.~Elhedhli, Z.~Li, and J.~H. Bookbinder.
\newblock Airfreight forwarding under system-wide and double discounts.
\newblock \emph{EURO Journal on Transportation and Logistics}, 6:\penalty0 165--183, 2017.

\bibitem[Fan and Li(2001)]{fan2001variable}
J.~Fan and R.~Li.
\newblock Variable selection via nonconcave penalized likelihood and its oracle properties.
\newblock \emph{Journal of the American Statistical Association}, 96\penalty0 (456):\penalty0 1348--1360, 2001.

\bibitem[Ge et~al.(2015)Ge, Huang, Jin, and Yuan]{ge2015escaping}
R.~Ge, F.~Huang, C.~Jin, and Y.~Yuan.
\newblock Escaping from saddle points--online stochastic gradient for tensor decomposition.
\newblock In \emph{Conference on Learning Theory}, pages 797--842, 2015.

\bibitem[Kingma and Ba(2015)]{kingma2015adam}
D.~P. Kingma and J.~Ba.
\newblock Adam: A method for stochastic optimization.
\newblock \emph{International Conference on Learning Representations (ICLR)}, 2015.

\bibitem[Martens(2010)]{martens2010deep}
J.~Martens.
\newblock Deep learning via hessian-free optimization.
\newblock In \emph{Proceedings of the 27th International Conference on Machine Learning (ICML-10)}, pages 735--742, 2010.

\bibitem[Nesterov and Polyak(2006)]{nesterov2006cubic}
Y.~Nesterov and B.~T. Polyak.
\newblock Cubic regularization of newton method and its global performance.
\newblock \emph{Mathematical Programming}, 108\penalty0 (1):\penalty0 177--205, 2006.

\bibitem[Nesterov(1983)]{nesterov1983method}
Y.~E. Nesterov.
\newblock A method for solving the convex programming problem with convergence rate o(1/k²).
\newblock \emph{Soviet Mathematics Doklady}, 27:\penalty0 372--376, 1983.

\bibitem[PEIRIS(2024)]{peiris2024transformative}
M.~S. PEIRIS.
\newblock Transformative integration of artificial intelligence in telemedicine, remote healthcare, and virtual patient monitoring: Enhancing diagnostic accuracy, personalizing care.
\newblock \emph{International Journal of Intelligent Healthcare Analytics}, 104\penalty0 (7):\penalty0 1019--1030, 2024.

\bibitem[Polyak(1964)]{polyak1964some}
B.~T. Polyak.
\newblock Some methods of speeding up the convergence of iteration methods.
\newblock \emph{USSR Computational Mathematics and Mathematical Physics}, 4\penalty0 (5):\penalty0 1--17, 1964.

\bibitem[Srivastava et~al.(2014)Srivastava, Hinton, Krizhevsky, Sutskever, and Salakhutdinov]{srivastava2014dropout}
N.~Srivastava, G.~Hinton, A.~Krizhevsky, I.~Sutskever, and R.~Salakhutdinov.
\newblock Dropout: A simple way to prevent neural networks from overfitting.
\newblock \emph{The Journal of Machine Learning Research}, 15\penalty0 (1):\penalty0 1929--1958, 2014.

\bibitem[Wang et~al.(2024)Wang, Zhu, Li, Wang, Qin, and Liu]{wang2024graph}
Z.~Wang, Y.~Zhu, Z.~Li, Z.~Wang, H.~Qin, and X.~Liu.
\newblock Graph neural network recommendation system for football formation.
\newblock \emph{Applied Science and Biotechnology Journal for Advanced Research}, 3\penalty0 (3):\penalty0 33--39, 2024.

\bibitem[Welling and Teh(2011)]{welling2011bayesian}
M.~Welling and Y.~W. Teh.
\newblock Bayesian learning via stochastic gradient langevin dynamics.
\newblock \emph{Proceedings of the 28th International Conference on Machine Learning (ICML-11)}, pages 681--688, 2011.

\bibitem[Xu et~al.(2012)Xu, Chang, Xu, and Zhang]{xu2012l12}
Z.~Xu, Y.~Chang, F.~Xu, and H.~Zhang.
\newblock L1/2 regularization: A thresholding representation theory and a fast solver.
\newblock \emph{IEEE Transactions on Neural Networks and Learning Systems}, 23\penalty0 (7):\penalty0 1013--1027, 2012.

\bibitem[Zhang(2010)]{zhang2010nearly}
C.-H. Zhang.
\newblock Nearly unbiased variable selection under minimax concave penalty.
\newblock \emph{The Annals of Statistics}, 38\penalty0 (2):\penalty0 894--942, 2010.

\end{thebibliography}

\end{document}